\documentclass[sigconf]{acmart}
\usepackage{makecell}
\usepackage{multirow}
\usepackage{booktabs}

\usepackage{graphicx}
\usepackage{amsmath}

\usepackage{amsfonts}
\usepackage{utfsym}
\usepackage{ulem}
\usepackage{float}
\usepackage[linesnumbered,ruled,vlined]{algorithm2e}
\AtBeginDocument{%
  }

\setcopyright{acmlicensed}
\copyrightyear{2024}
\acmYear{2024}
\acmDOI{XXXXXXX.XXXXXXX}

\acmConference[MM ’24]{the 32st ACM International Conference on Multimedia }{28 October - 1 November 2024}{Melbourne, Australia}
\acmISBN{978-1-4503-XXXX-X/18/06}

\begin{document}

\title{The Solution for the 5th GCAIAC Zero-shot Referring Expression Comprehension Challenge}

\author{
Longfei Huang$^{1}$ , Feng Yu$^{1}$ , Zhihao Guan$^{1}$ , Zhonghua Wan$^{1}$ , Yang Yang$^{1}$ 
\and
$^1$ Nanjing University of Science and Technology \\
}
\renewcommand{\shortauthors}{Trovato et al.}

\begin{abstract}
  This report presents a solution for the zero-shot referring expression comprehension task. Visual-language multimodal base models (such as CLIP, SAM) have gained significant attention in recent years as a cornerstone of mainstream research. One of the key applications of multimodal base models lies in their ability to generalize to zero-shot downstream tasks. Unlike traditional referring expression comprehension, zero-shot referring expression comprehension aims to apply pre-trained visual-language models directly to the task without specific training. Recent studies have enhanced the zero-shot performance of multimodal base models in referring expression comprehension tasks by introducing visual prompts. To address the zero-shot referring expression comprehension challenge, we introduced a combination of visual prompts and considered the influence of textual prompts, employing joint prediction tailored to the data characteristics. Ultimately, our approach achieved accuracy rates of 84.825 on the A leaderboard and 71.460 on the B leaderboard, securing the first position.
\end{abstract}




\maketitle
\setlength{\intextsep}{2pt}
\setlength{\abovecaptionskip}{1pt}

\begin{figure}[h]
  \centering
  \includegraphics[trim=0 0 0 0, clip, width=0.48\textwidth]{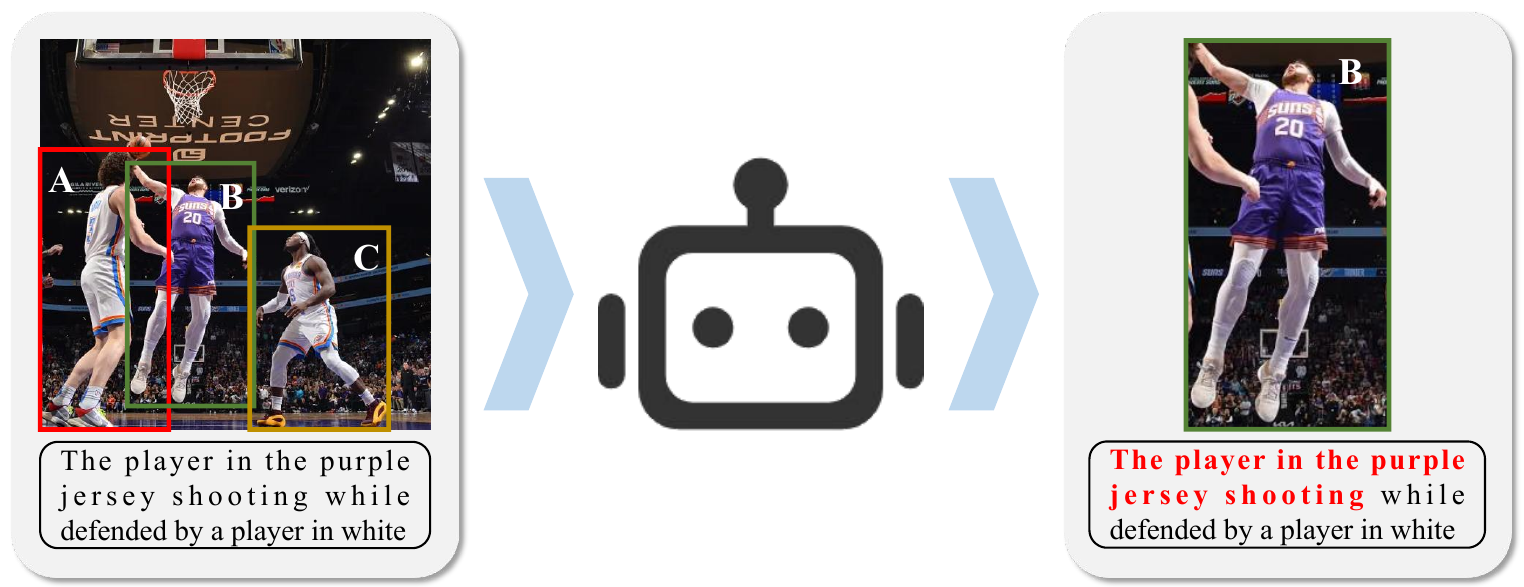}
  \caption{Referring Expression Comprehension aims to locate the position of a specific object in an image based on a provided textual description, such as accurately locating the proposal of the player in a purple jersey shooting skillfully in the image.}
  \label{fig: Rec_example}
\end{figure}

\section{Introduction}
With the advancement and application of multimodal research \cite{YangGLLLY24, YangBGZYY23, abs-2110-11767, YangYBZZGXY23, YangZWLXJ21}, large-scale Vision-Language Models (VLMs) have demonstrated remarkable capabilities in visual-related tasks and applications \cite{BowmanAZTPPS23, SchaefferMK23}. Leveraging abundant data and extensive model capacity, these models exhibit the potential for transfer to various downstream tasks, including visual grounding \cite{ZareianRHC21}, image generation \cite{LiLWMYGLL23}, and visual reasoning \cite{ZellersBFC19}. Typically, VLMs require additional data for fine-tuning to adapt to specific downstream tasks. However, further research is needed to explore directly utilizing pre-trained VLMs for their intrinsic abilities in visual-textual comprehension within downstream tasks.

Zero-shot Referring Expression Comprehension tasks \cite{SubramanianMD0022, Shtedritski0V23, YangWLWY23} aim to leverage existing large-scale multimodal models, such as CLIP \cite{RadfordKHRGASAM21}. However, these foundational multimodal models are insensitive to precise spatial relationships like object localization \cite{SubramanianMD0022}. Achieving high-performance zero-shot referring expression comprehension (REC) tasks involves designing effective visual prompts and other reasoning strategies. REC tasks involve locating target objects in images based on linguistic descriptions, as illustrated in Figure \ref{fig: Rec_example}.

\cite{Shtedritski0V23} indicates that Visual Language Models (VLMs) may inherently possess the capability to comprehend local objects within entire images. Therefore, specially designed visual prompts have the potential to explicitly invoke this ability of VLMs, thereby benefiting various tasks. Visual prompts employ distinct markers (such as boxes or circles) placed over relevant regions in images to effectively guide the attention of VLMs. Currently, many studies are leveraging this capability by designing a variety of unique visual prompts to ensure that models focus on highlighted relevant information while retaining the overall contextual details of the image.

However, existing methods \cite{SubramanianMD0022, Shtedritski0V23, YangWLWY23} predominantly focus on designing fine-grained visual prompts to activate VLMs' ability to understand object spatial localization, while neglecting the impact of coarse-grained visual enhancements and textual factors. To address this issue and maximize the zero-shot performance of multimodal large-scale models while considering their reasoning capabilities, we propose a strategy involving visual prompt combinations and text redundancy reduction operations. Additionally, we introduce a method for joint prediction tailored to the characteristics of the data.

\section{Related Work}
\subsection{Vision-Language Models} 
Large-scale language models (LLMs) can solve complex problems with minimal data through fine-tuning and exhibit remarkable zero-shot transferability \cite{SchaefferMK23}. In recent years, with the advancement and application of multimodal models \cite{0074ZGGZ22, YangZXYZY21, YangWZX019, YangZWLXJ21, YangFZLJ21}, Vision-Language Models (VLMs) trained on image-text data sourced from the web have shown outstanding performance in computer vision (CV) tasks, demonstrating commendable zero-shot capabilities \cite{RadfordKHRGASAM21}. Models like CLIP \cite{RadfordKHRGASAM21} and UNITER \cite{ChenLYK0G0020}, trained via contrastive learning, have excelled in tasks such as image classification. However, instance-level tasks such as referring expression comprehension often require adjustments to existing methods in visual and textual encoders, as well as image grounding techniques. In contrast, this paper proposes a zero-shot architecture for instance-level tasks using pre-trained VLMs.

\subsection{Prompt Engineering}
Prompt engineering significantly enhances the emerging capabilities of LLMs in natural language processing. Common types of prompt inputs primarily involve adding pre-designed fixed prompts or a set of learnable tokens at the beginning of textual inputs \cite{KojimaGRMI22, Li_2024_CVPR}. Prompting has since expanded to include open vocabulary detection and segmentation. While language prompting has been extensively explored, visual prompting has received less attention. Early research utilized visual prompts to adapt VLMs. Recently, \cite{YangWLWY23} achieved fine-grained visual prompting through the SAM model \cite{Kirillov_2023_ICCV}, enhancing the performance of the CLIP model \cite{RadfordKHRGASAM21}.

However, existing methods \cite{SubramanianMD0022, Shtedritski0V23, YangWLWY23} have predominantly considered visual prompting at a fine-grained level, overlooking the respective strengths and weaknesses of coarse-grained and fine-grained approaches. They have not addressed the synergy between coarse and fine-grained visual enhancements nor the role of textual prompts. Therefore, we propose a method involving visual prompt combinations and text redundancy reduction.

\subsection{Referring Expression Comprehension} 

Referring Expression Comprehension (REC) tasks \cite{CaoJCZ22} involve locating the bounding box of a target object in an image based on linguistic descriptions. Mainstream REC methods can generally be categorized into two types: proposal-based methods and proposal-free methods. Proposal-based methods utilize object detectors (such as Mask R-CNN \cite{RenHG017}) to detect instance regions, which are then cropped and used for subsequent classification. Proposal-free methods (e.g., MDETR \cite{KamathSLSMC21}) involve end-to-end training of vision-language models. Recently, zero-shot approaches have integrated proposal boxes from MAttNet \cite{Yu0SYLBB18} with Vision-Language models for tasks like image caption matching.

\begin{figure*}[t]
  \centering
  \includegraphics[trim=0 0 0 0, clip, width=0.82\textwidth]{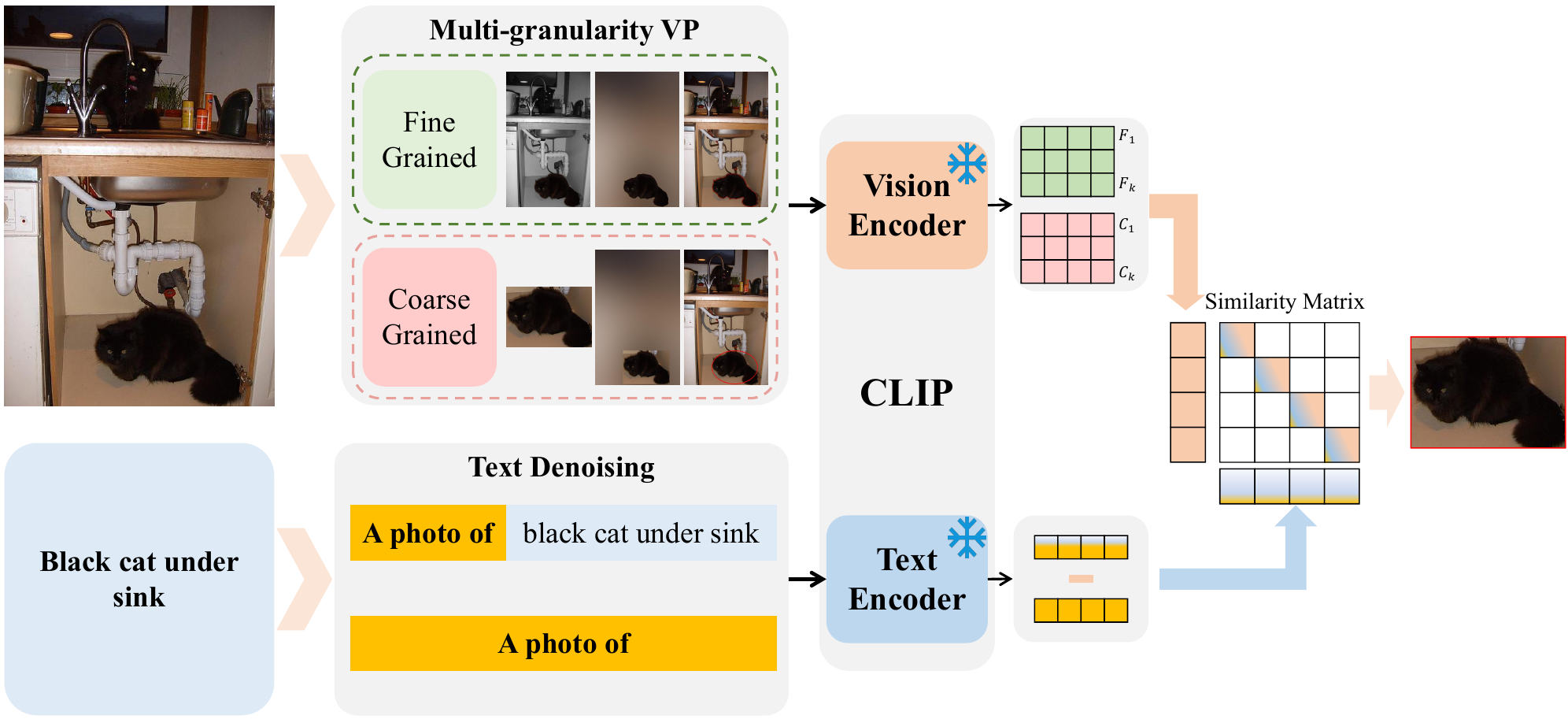}
  \caption{Overall Architecture. We use visual prompts to mark regions of interest in images and simultaneously process textual content to enhance the visual-language comprehension capabilities of multimodal models.}
  \label{fig: framework}
\end{figure*}

\section{Method}
Our approach consists of three main components: visual prompts, text redundancy reduction, and joint prediction. The main framework of our method is illustrated in Figure \ref{fig: framework}.

\subsection{Visual Prompt}
\begin{figure*}[t]
  \centering
  \includegraphics[trim=0 0 0 0, clip, width=0.80\textwidth]{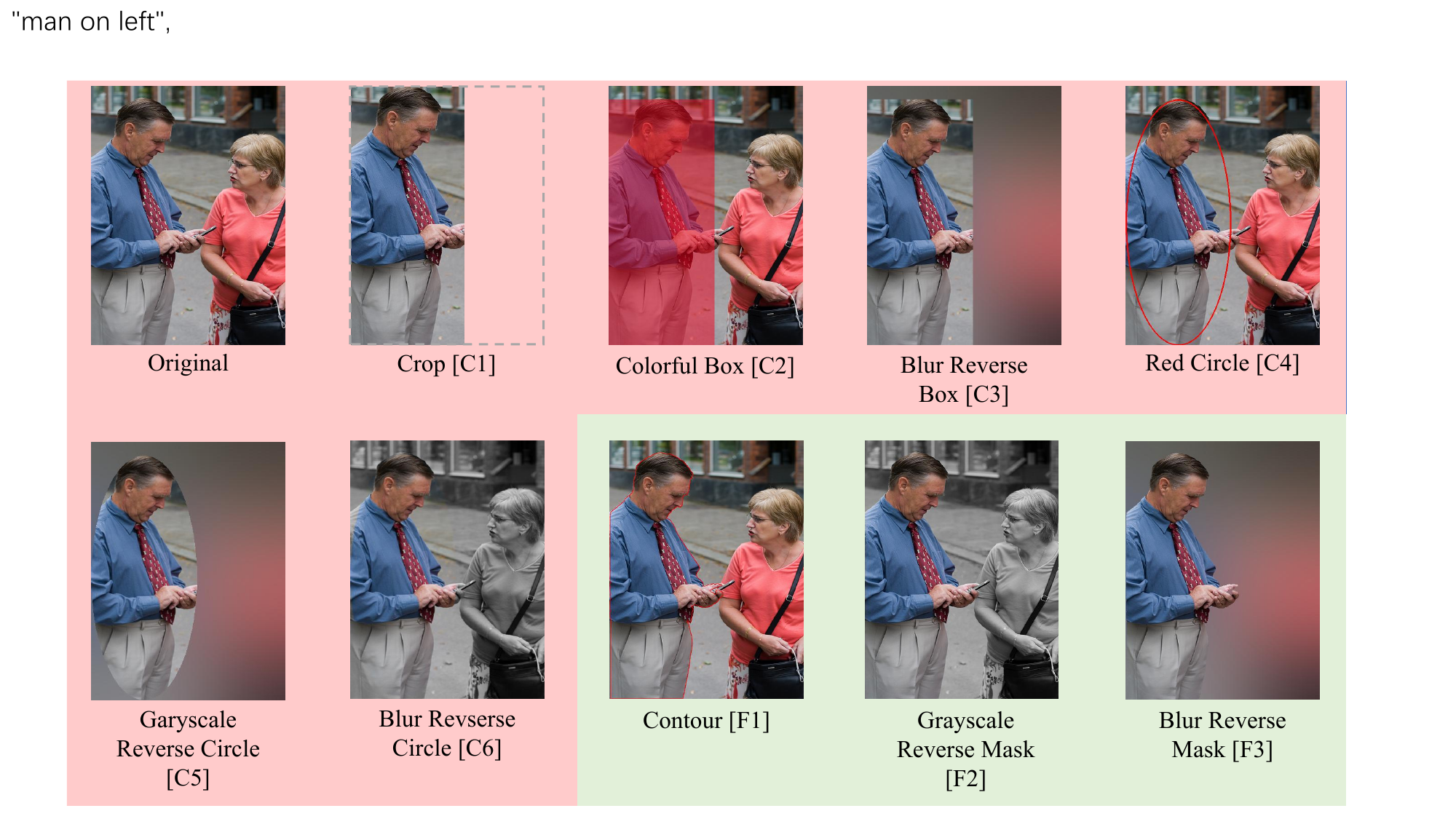}
  \caption{A summary of the multi-granularity vision prompts used in this paper with the caption "man on left". Note that vision prompts highlighted in pink represent coarse-grained vision prompts, while those highlighted in green represent fine-grained vision prompts. The main difference between coarse-grained and fine-grained vision prompt is the use of SAM to segment objects within the proposal, followed by employing box or circle-based methods to highlight or mask the target object.}
  \label{fig: VP_summary}
\end{figure*}

In terms of visual prompts, we categorize them into coarse-grained and fine-grained types, as illustrated in Figure \ref{fig: VP_summary}. We explore various visual prompt configurations. Unlike approaches focusing exclusively on fine-grained visual prompts, we utilize a combination of coarse-grained and fine-grained visual prompts. This approach ensures that the model maintains detection performance for fine-grained objects while considering spatial relationships among targets, thereby further activating the VLM's visual-language comprehension abilities. Through experiments, we determined that the optimal visual prompt combination is C1, C3, C4, F1, F2, F3.

\subsection{Removing Redundant Text}
We draw inspiration from NLP techniques for removing redundant text content to enhance the visual-language comprehension capabilities of CLIP. By eliminating redundant portions of text prompts within CLIP, we aim to improve its performance in image-text matching tasks. Adding a formatted prompt like "a photo of" before text inputs have been shown to effectively enhance CLIP's performance, but these added prompts themselves do not contribute meaningful information and are considered redundant for the textual context. By performing redundancy reduction operations, we further boost CLIP's effectiveness.

This approach is akin to the Subtraction method proposed by RedCircle \cite{Shtedritski0V23}. Subtraction leverages a large volume of negative sample text to identify noise within texts. By removing this noise from positive samples, interference is reduced, thus improving model performance. However, Subtraction requires extensive negative sample texts, making it computationally intensive. Moreover, similarities in semantic content between negative and positive samples may lead to performance degradation if improperly managed, rendering the denoising effect uncontrollable.

In contrast, our text redundancy reduction method employs a more direct approach, akin to hard denoising. It removes overtly noisy prompt words from the text while preserving original information. This significantly reduces computational overhead while offering greater control over denoising effects compared to soft denoising methods like Subtraction. Experimental results demonstrate that our approach achieves superior performance.

\subsection{Joint Prediction}

To address the strong prior knowledge where each image corresponds to multiple entries, with each entry containing several textual descriptions pointing to the same bounding box, and different entries pointing to different bounding boxes within the same image, we propose a method called joint prediction.

For different descriptions within the same entry, we aggregate all prediction results and use them as a collective prediction for all descriptions. For different entries within the same image, we employ the Hungarian algorithm to find the optimal solution, ensuring distinct predictions for different entries.

\section{Experiments}

\subsection{Experiments Setup}
\textbf{Dataset and Model}. In the zero-shot setting, inference is conducted solely on the official test set provided, without any training data available. The model exclusively uses the CLIP model for inference and is augmented with the SAM model to assist in inference tasks. 

\textbf{Implementation Details}. We utilize the CLIP trained by OpenAI, namely ViT-B/32 and RN50×16 backbones. For the fine-grained visual prompt, we employ SAM-ViT-H, a variant of Segment Anything Model. All experiments are conducted on a single RTX 3090 GPU. More implementation details are presented in the supplementary materials. Redcircle has a line thickness of 6 pixels, while the target outline has a thickness of 2 pixels. The image is padded using gray color (100, 100, 100), and Gaussian blur is applied with a standard deviation of 100. The candidate box is constrained to be greater than or equal to 0.05 of the image area.

\subsection{Main Results}
Our method achieved an accuracy of 72.968 on the offline test set, 84.825 on the A leaderboard, and 71.46 on the B leaderboard. The ablation results of our method on the offline test are shown in Table \ref{tab:table-1}. Experimental results indicate that our approach, combining visual prompt combinations and text redundancy reduction, effectively enhances CLIP's visual-language comprehension capabilities.

\begin{table}[!ht]
\centering
\begin{tabular}{ccc}
\toprule[1.5pt]
\# & \textbf{Method} &  \textbf{ACC} \\ 
\midrule 
1 & ReCLIP & 46.79 \\
2 & +visual prompt & 53.069 \\
3 & +removing redundant & 60.846 \\
4 & +parameter tuning & 63.389 \\
5 & +joint prediction & 72.968 \\
\bottomrule[1.5pt]
\end{tabular}
\caption{Ablation experiments.} \label{tab:table-1}
\end{table}

\section{Conclusion}
In this paper, we enhance the visual-language comprehension of Vision-Language Models (VLMs) at the visual level by employing a combination of visual prompt methods, incorporating finer-grained prompts using the SAM model alongside coarse-grained prompts. Additionally, we improve the performance of the CLIP model by optimizing textual prompts through redundancy reduction, which proves more efficient and effective compared to prevalent negative sample denoising methods. Furthermore, we optimize predictions through joint prediction to further enhance model performance. Experimental results demonstrate that our approach significantly enhances the visual-language comprehension capabilities of multimodal base models and exhibits robustness.


\bibliographystyle{ACM-Reference-Format}
\bibliography{main}


\begin{thebibliography}{26}


\ifx \showCODEN    \undefined \def \showCODEN     #1{\unskip}     \fi
\ifx \showDOI      \undefined \def \showDOI       #1{#1}\fi
\ifx \showISBNx    \undefined \def \showISBNx     #1{\unskip}     \fi
\ifx \showISBNxiii \undefined \def \showISBNxiii  #1{\unskip}     \fi
\ifx \showISSN     \undefined \def \showISSN      #1{\unskip}     \fi
\ifx \showLCCN     \undefined \def \showLCCN      #1{\unskip}     \fi
\ifx \shownote     \undefined \def \shownote      #1{#1}          \fi
\ifx \showarticletitle \undefined \def \showarticletitle #1{#1}   \fi
\ifx \showURL      \undefined \def \showURL       {\relax}        \fi
\providecommand\bibfield[2]{#2}
\providecommand\bibinfo[2]{#2}
\providecommand\natexlab[1]{#1}
\providecommand\showeprint[2][]{arXiv:#2}

\bibitem[Bowman et~al\mbox{.}(2023)]%
        {BowmanAZTPPS23}
\bibfield{author}{\bibinfo{person}{Benjamin Bowman}, \bibinfo{person}{Alessandro Achille}, \bibinfo{person}{Luca Zancato}, \bibinfo{person}{Matthew Trager}, \bibinfo{person}{Pramuditha Perera}, \bibinfo{person}{Giovanni Paolini}, {and} \bibinfo{person}{Stefano Soatto}.} \bibinfo{year}{2023}\natexlab{}.
\newblock \showarticletitle{{\`{A}}-la-carte Prompt Tuning {(APT):} Combining Distinct Data Via Composable Prompting}. In \bibinfo{booktitle}{\emph{CVPR}}. \bibinfo{publisher}{{IEEE}}, \bibinfo{pages}{14984--14993}.
\newblock


\bibitem[Cao et~al\mbox{.}(2022)]%
        {CaoJCZ22}
\bibfield{author}{\bibinfo{person}{Meng Cao}, \bibinfo{person}{Ji Jiang}, \bibinfo{person}{Long Chen}, {and} \bibinfo{person}{Yuexian Zou}.} \bibinfo{year}{2022}\natexlab{}.
\newblock \showarticletitle{Correspondence Matters for Video Referring Expression Comprehension}. In \bibinfo{booktitle}{\emph{{MM} '22: The 30th {ACM} International Conference on Multimedia, Lisboa, Portugal, October 10 - 14, 2022}}. \bibinfo{publisher}{{ACM}}, \bibinfo{pages}{4967--4976}.
\newblock


\bibitem[Chen et~al\mbox{.}(2020)]%
        {ChenLYK0G0020}
\bibfield{author}{\bibinfo{person}{Yen{-}Chun Chen}, \bibinfo{person}{Linjie Li}, \bibinfo{person}{Licheng Yu}, \bibinfo{person}{Ahmed~El Kholy}, \bibinfo{person}{Faisal Ahmed}, \bibinfo{person}{Zhe Gan}, \bibinfo{person}{Yu Cheng}, {and} \bibinfo{person}{Jingjing Liu}.} \bibinfo{year}{2020}\natexlab{}.
\newblock \showarticletitle{{UNITER:} UNiversal Image-TExt Representation Learning}. In \bibinfo{booktitle}{\emph{ECCV}} \emph{(\bibinfo{series}{Lecture Notes in Computer Science}, Vol.~\bibinfo{volume}{12375})}. \bibinfo{publisher}{Springer}, \bibinfo{pages}{104--120}.
\newblock


\bibitem[Kamath et~al\mbox{.}(2021)]%
        {KamathSLSMC21}
\bibfield{author}{\bibinfo{person}{Aishwarya Kamath}, \bibinfo{person}{Mannat Singh}, \bibinfo{person}{Yann LeCun}, \bibinfo{person}{Gabriel Synnaeve}, \bibinfo{person}{Ishan Misra}, {and} \bibinfo{person}{Nicolas Carion}.} \bibinfo{year}{2021}\natexlab{}.
\newblock \showarticletitle{{MDETR} - Modulated Detection for End-to-End Multi-Modal Understanding}. In \bibinfo{booktitle}{\emph{ICCV}}. \bibinfo{publisher}{{IEEE}}, \bibinfo{pages}{1760--1770}.
\newblock


\bibitem[Kirillov et~al\mbox{.}(2023)]%
        {Kirillov_2023_ICCV}
\bibfield{author}{\bibinfo{person}{Alexander Kirillov}, \bibinfo{person}{Eric Mintun}, \bibinfo{person}{Nikhila Ravi}, \bibinfo{person}{Hanzi Mao}, \bibinfo{person}{Chloe Rolland}, \bibinfo{person}{Laura Gustafson}, \bibinfo{person}{Tete Xiao}, \bibinfo{person}{Spencer Whitehead}, \bibinfo{person}{Alexander~C. Berg}, \bibinfo{person}{Wan-Yen Lo}, \bibinfo{person}{Piotr Dollar}, {and} \bibinfo{person}{Ross Girshick}.} \bibinfo{year}{2023}\natexlab{}.
\newblock \showarticletitle{Segment Anything}. In \bibinfo{booktitle}{\emph{ICCV}}. \bibinfo{pages}{4015--4026}.
\newblock


\bibitem[Kojima et~al\mbox{.}(2022)]%
        {KojimaGRMI22}
\bibfield{author}{\bibinfo{person}{Takeshi Kojima}, \bibinfo{person}{Shixiang~Shane Gu}, \bibinfo{person}{Machel Reid}, \bibinfo{person}{Yutaka Matsuo}, {and} \bibinfo{person}{Yusuke Iwasawa}.} \bibinfo{year}{2022}\natexlab{}.
\newblock \showarticletitle{Large Language Models are Zero-Shot Reasoners}. In \bibinfo{booktitle}{\emph{NeurIPS}}.
\newblock


\bibitem[Li et~al\mbox{.}(2024)]%
        {Li_2024_CVPR}
\bibfield{author}{\bibinfo{person}{Lin Li}, \bibinfo{person}{Haoyan Guan}, \bibinfo{person}{Jianing Qiu}, {and} \bibinfo{person}{Michael Spratling}.} \bibinfo{year}{2024}\natexlab{}.
\newblock \showarticletitle{One Prompt Word is Enough to Boost Adversarial Robustness for Pre-trained Vision-Language Models}. In \bibinfo{booktitle}{\emph{CVPR}}. \bibinfo{pages}{24408--24419}.
\newblock


\bibitem[Li et~al\mbox{.}(2023)]%
        {LiLWMYGLL23}
\bibfield{author}{\bibinfo{person}{Yuheng Li}, \bibinfo{person}{Haotian Liu}, \bibinfo{person}{Qingyang Wu}, \bibinfo{person}{Fangzhou Mu}, \bibinfo{person}{Jianwei Yang}, \bibinfo{person}{Jianfeng Gao}, \bibinfo{person}{Chunyuan Li}, {and} \bibinfo{person}{Yong~Jae Lee}.} \bibinfo{year}{2023}\natexlab{}.
\newblock \showarticletitle{{GLIGEN:} Open-Set Grounded Text-to-Image Generation}. In \bibinfo{booktitle}{\emph{CVPR}}. \bibinfo{publisher}{{IEEE}}, \bibinfo{pages}{22511--22521}.
\newblock


\bibitem[Radford et~al\mbox{.}(2021)]%
        {RadfordKHRGASAM21}
\bibfield{author}{\bibinfo{person}{Alec Radford}, \bibinfo{person}{Jong~Wook Kim}, \bibinfo{person}{Chris Hallacy}, \bibinfo{person}{Aditya Ramesh}, \bibinfo{person}{Gabriel Goh}, \bibinfo{person}{Sandhini Agarwal}, \bibinfo{person}{Girish Sastry}, \bibinfo{person}{Amanda Askell}, \bibinfo{person}{Pamela Mishkin}, \bibinfo{person}{Jack Clark}, \bibinfo{person}{Gretchen Krueger}, {and} \bibinfo{person}{Ilya Sutskever}.} \bibinfo{year}{2021}\natexlab{}.
\newblock \showarticletitle{Learning Transferable Visual Models From Natural Language Supervision}. In \bibinfo{booktitle}{\emph{ICML}} \emph{(\bibinfo{series}{Proceedings of Machine Learning Research}, Vol.~\bibinfo{volume}{139})}. \bibinfo{publisher}{{PMLR}}, \bibinfo{pages}{8748--8763}.
\newblock


\bibitem[Ren et~al\mbox{.}(2017)]%
        {RenHG017}
\bibfield{author}{\bibinfo{person}{Shaoqing Ren}, \bibinfo{person}{Kaiming He}, \bibinfo{person}{Ross~B. Girshick}, {and} \bibinfo{person}{Jian Sun}.} \bibinfo{year}{2017}\natexlab{}.
\newblock \showarticletitle{Faster {R-CNN:} Towards Real-Time Object Detection with Region Proposal Networks}.
\newblock \bibinfo{journal}{\emph{{IEEE} Trans. Pattern Anal. Mach. Intell.}} \bibinfo{volume}{39}, \bibinfo{number}{6} (\bibinfo{year}{2017}), \bibinfo{pages}{1137--1149}.
\newblock


\bibitem[Schaeffer et~al\mbox{.}(2023)]%
        {SchaefferMK23}
\bibfield{author}{\bibinfo{person}{Rylan Schaeffer}, \bibinfo{person}{Brando Miranda}, {and} \bibinfo{person}{Sanmi Koyejo}.} \bibinfo{year}{2023}\natexlab{}.
\newblock \showarticletitle{Are Emergent Abilities of Large Language Models a Mirage?}. In \bibinfo{booktitle}{\emph{NeurIPS}}.
\newblock


\bibitem[Shtedritski et~al\mbox{.}(2023)]%
        {Shtedritski0V23}
\bibfield{author}{\bibinfo{person}{Aleksandar Shtedritski}, \bibinfo{person}{Christian Rupprecht}, {and} \bibinfo{person}{Andrea Vedaldi}.} \bibinfo{year}{2023}\natexlab{}.
\newblock \showarticletitle{What does {CLIP} know about a red circle? Visual prompt engineering for VLMs}. In \bibinfo{booktitle}{\emph{ICCV}}. \bibinfo{publisher}{{IEEE}}, \bibinfo{pages}{11953--11963}.
\newblock


\bibitem[Subramanian et~al\mbox{.}(2022)]%
        {SubramanianMD0022}
\bibfield{author}{\bibinfo{person}{Sanjay Subramanian}, \bibinfo{person}{William Merrill}, \bibinfo{person}{Trevor Darrell}, \bibinfo{person}{Matt Gardner}, \bibinfo{person}{Sameer Singh}, {and} \bibinfo{person}{Anna Rohrbach}.} \bibinfo{year}{2022}\natexlab{}.
\newblock \showarticletitle{ReCLIP: {A} Strong Zero-Shot Baseline for Referring Expression Comprehension}. In \bibinfo{booktitle}{\emph{ACL}}. \bibinfo{publisher}{Association for Computational Linguistics}, \bibinfo{pages}{5198--5215}.
\newblock


\bibitem[Yang et~al\mbox{.}(2023b)]%
        {YangWLWY23}
\bibfield{author}{\bibinfo{person}{Lingfeng Yang}, \bibinfo{person}{Yueze Wang}, \bibinfo{person}{Xiang Li}, \bibinfo{person}{Xinlong Wang}, {and} \bibinfo{person}{Jian Yang}.} \bibinfo{year}{2023}\natexlab{b}.
\newblock \showarticletitle{Fine-Grained Visual Prompting}. In \bibinfo{booktitle}{\emph{Advances in Neural Information Processing Systems 36: Annual Conference on Neural Information Processing Systems 2023, NeurIPS 2023, New Orleans, LA, USA, December 10 - 16, 2023}}.
\newblock


\bibitem[Yang et~al\mbox{.}(2023a)]%
        {YangBGZYY23}
\bibfield{author}{\bibinfo{person}{Yang Yang}, \bibinfo{person}{Ran Bao}, \bibinfo{person}{Weili Guo}, \bibinfo{person}{De{-}Chuan Zhan}, \bibinfo{person}{Yilong Yin}, {and} \bibinfo{person}{Jian Yang}.} \bibinfo{year}{2023}\natexlab{a}.
\newblock \showarticletitle{Deep visual-linguistic fusion network considering cross-modal inconsistency for rumor detection}.
\newblock \bibinfo{journal}{\emph{Sci. China Inf. Sci.}} \bibinfo{volume}{66}, \bibinfo{number}{12} (\bibinfo{year}{2023}).
\newblock


\bibitem[Yang et~al\mbox{.}(2021a)]%
        {YangFZLJ21}
\bibfield{author}{\bibinfo{person}{Yang Yang}, \bibinfo{person}{Zhao{-}Yang Fu}, \bibinfo{person}{De{-}Chuan Zhan}, \bibinfo{person}{Zhi{-}Bin Liu}, {and} \bibinfo{person}{Yuan Jiang}.} \bibinfo{year}{2021}\natexlab{a}.
\newblock \showarticletitle{Semi-Supervised Multi-Modal Multi-Instance Multi-Label Deep Network with Optimal Transport}.
\newblock \bibinfo{journal}{\emph{{IEEE} Trans. Knowl. Data Eng.}} \bibinfo{volume}{33}, \bibinfo{number}{2} (\bibinfo{year}{2021}), \bibinfo{pages}{696--709}.
\newblock


\bibitem[Yang et~al\mbox{.}(2024)]%
        {YangGLLLY24}
\bibfield{author}{\bibinfo{person}{Yang Yang}, \bibinfo{person}{Jinyi Guo}, \bibinfo{person}{Guangyu Li}, \bibinfo{person}{Lanyu Li}, \bibinfo{person}{Wenjie Li}, {and} \bibinfo{person}{Jian Yang}.} \bibinfo{year}{2024}\natexlab{}.
\newblock \showarticletitle{Alignment efficient image-sentence retrieval considering transferable cross-modal representation learning}.
\newblock \bibinfo{journal}{\emph{Frontiers Comput. Sci.}} \bibinfo{volume}{18}, \bibinfo{number}{3} (\bibinfo{year}{2024}), \bibinfo{pages}{181335}.
\newblock


\bibitem[Yang et~al\mbox{.}(2019)]%
        {YangWZX019}
\bibfield{author}{\bibinfo{person}{Yang Yang}, \bibinfo{person}{Ke{-}Tao Wang}, \bibinfo{person}{De{-}Chuan Zhan}, \bibinfo{person}{Hui Xiong}, {and} \bibinfo{person}{Yuan Jiang}.} \bibinfo{year}{2019}\natexlab{}.
\newblock \showarticletitle{Comprehensive Semi-Supervised Multi-Modal Learning}. In \bibinfo{booktitle}{\emph{IJCAI}}. \bibinfo{publisher}{ijcai.org}, \bibinfo{pages}{4092--4098}.
\newblock


\bibitem[Yang et~al\mbox{.}(2021b)]%
        {abs-2110-11767}
\bibfield{author}{\bibinfo{person}{Yang Yang}, \bibinfo{person}{Hongchen Wei}, \bibinfo{person}{Hengshu Zhu}, \bibinfo{person}{Dianhai Yu}, \bibinfo{person}{Hui Xiong}, {and} \bibinfo{person}{Jian Yang}.} \bibinfo{year}{2021}\natexlab{b}.
\newblock \showarticletitle{Exploiting Cross-Modal Prediction and Relation Consistency for Semi-Supervised Image Captioning}.
\newblock \bibinfo{journal}{\emph{CoRR}}  \bibinfo{volume}{abs/2110.11767} (\bibinfo{year}{2021}).
\newblock


\bibitem[Yang et~al\mbox{.}(2023c)]%
        {YangYBZZGXY23}
\bibfield{author}{\bibinfo{person}{Yang Yang}, \bibinfo{person}{Jia{-}Qi Yang}, \bibinfo{person}{Ran Bao}, \bibinfo{person}{De{-}Chuan Zhan}, \bibinfo{person}{Hengshu Zhu}, \bibinfo{person}{Xiaoru Gao}, \bibinfo{person}{Hui Xiong}, {and} \bibinfo{person}{Jian Yang}.} \bibinfo{year}{2023}\natexlab{c}.
\newblock \showarticletitle{Corporate Relative Valuation Using Heterogeneous Multi-Modal Graph Neural Network}.
\newblock \bibinfo{journal}{\emph{{IEEE} Trans. Knowl. Data Eng.}} \bibinfo{volume}{35}, \bibinfo{number}{1} (\bibinfo{year}{2023}), \bibinfo{pages}{211--224}.
\newblock


\bibitem[Yang et~al\mbox{.}(2021c)]%
        {YangZWLXJ21}
\bibfield{author}{\bibinfo{person}{Yang Yang}, \bibinfo{person}{De{-}Chuan Zhan}, \bibinfo{person}{Yi{-}Feng Wu}, \bibinfo{person}{Zhi{-}Bin Liu}, \bibinfo{person}{Hui Xiong}, {and} \bibinfo{person}{Yuan Jiang}.} \bibinfo{year}{2021}\natexlab{c}.
\newblock \showarticletitle{Semi-Supervised Multi-Modal Clustering and Classification with Incomplete Modalities}.
\newblock \bibinfo{journal}{\emph{{IEEE} Trans. Knowl. Data Eng.}} \bibinfo{volume}{33}, \bibinfo{number}{2} (\bibinfo{year}{2021}), \bibinfo{pages}{682--695}.
\newblock


\bibitem[Yang et~al\mbox{.}(2021d)]%
        {YangZXYZY21}
\bibfield{author}{\bibinfo{person}{Yang Yang}, \bibinfo{person}{Chubing Zhang}, \bibinfo{person}{Yi{-}Chu Xu}, \bibinfo{person}{Dianhai Yu}, \bibinfo{person}{De{-}Chuan Zhan}, {and} \bibinfo{person}{Jian Yang}.} \bibinfo{year}{2021}\natexlab{d}.
\newblock \showarticletitle{Rethinking Label-Wise Cross-Modal Retrieval from {A} Semantic Sharing Perspective}. In \bibinfo{booktitle}{\emph{IJCAI}}. \bibinfo{publisher}{ijcai.org}, \bibinfo{pages}{3300--3306}.
\newblock


\bibitem[Yang et~al\mbox{.}(2022)]%
        {0074ZGGZ22}
\bibfield{author}{\bibinfo{person}{Yang Yang}, \bibinfo{person}{Jingshuai Zhang}, \bibinfo{person}{Fan Gao}, \bibinfo{person}{Xiaoru Gao}, {and} \bibinfo{person}{Hengshu Zhu}.} \bibinfo{year}{2022}\natexlab{}.
\newblock \showarticletitle{{DOMFN:} {A} Divergence-Orientated Multi-Modal Fusion Network for Resume Assessment}. In \bibinfo{booktitle}{\emph{{MM} '22: The 30th {ACM} International Conference on Multimedia, Lisboa, Portugal, October 10 - 14, 2022}}. \bibinfo{publisher}{{ACM}}, \bibinfo{pages}{1612--1620}.
\newblock


\bibitem[Yu et~al\mbox{.}(2018)]%
        {Yu0SYLBB18}
\bibfield{author}{\bibinfo{person}{Licheng Yu}, \bibinfo{person}{Zhe Lin}, \bibinfo{person}{Xiaohui Shen}, \bibinfo{person}{Jimei Yang}, \bibinfo{person}{Xin Lu}, \bibinfo{person}{Mohit Bansal}, {and} \bibinfo{person}{Tamara~L. Berg}.} \bibinfo{year}{2018}\natexlab{}.
\newblock \showarticletitle{MAttNet: Modular Attention Network for Referring Expression Comprehension}. In \bibinfo{booktitle}{\emph{CVPR}}. \bibinfo{publisher}{Computer Vision Foundation / {IEEE} Computer Society}, \bibinfo{pages}{1307--1315}.
\newblock


\bibitem[Zareian et~al\mbox{.}(2021)]%
        {ZareianRHC21}
\bibfield{author}{\bibinfo{person}{Alireza Zareian}, \bibinfo{person}{Kevin~Dela Rosa}, \bibinfo{person}{Derek~Hao Hu}, {and} \bibinfo{person}{Shih{-}Fu Chang}.} \bibinfo{year}{2021}\natexlab{}.
\newblock \showarticletitle{Open-Vocabulary Object Detection Using Captions}. In \bibinfo{booktitle}{\emph{{IEEE} Conference on Computer Vision and Pattern Recognition, {CVPR} 2021, virtual, June 19-25, 2021}}. \bibinfo{publisher}{Computer Vision Foundation / {IEEE}}, \bibinfo{pages}{14393--14402}.
\newblock


\bibitem[Zellers et~al\mbox{.}(2019)]%
        {ZellersBFC19}
\bibfield{author}{\bibinfo{person}{Rowan Zellers}, \bibinfo{person}{Yonatan Bisk}, \bibinfo{person}{Ali Farhadi}, {and} \bibinfo{person}{Yejin Choi}.} \bibinfo{year}{2019}\natexlab{}.
\newblock \showarticletitle{From Recognition to Cognition: Visual Commonsense Reasoning}. In \bibinfo{booktitle}{\emph{CVPR}}. \bibinfo{publisher}{Computer Vision Foundation / {IEEE}}, \bibinfo{pages}{6720--6731}.
\newblock


\end{thebibliography}





\end{document}